\documentclass[pdflatex,sn-mathphys]{sn-jnl}

\usepackage{pifont}
\usepackage{soul}

\jyear{2022}%

\theoremstyle{thmstyleone}%
\theoremstyle{thmstyletwo}%

\theoremstyle{thmstylethree}%

\newcommand{\totalvids}{28}
\newcommand{\totalclips}{76}
\newcommand{\totalanno}{8.1k}
\newcommand{\totalannotations}{8,178}
\newcommand{\ourdataset}{\textit{Surgical Hands}}
\newcommand{\ourdnn}{\textbf{CondPose}}

\begin{document}

\title[Temporally Guided Articulated Hand Pose Tracking in Surgical Videos]{Temporally Guided Articulated Hand Pose \\Tracking in Surgical Videos}


\author*[1]{\fnm{Nathan} \sur{Louis}}\email{natlouis@umich.edu}
\author[2]{\fnm{Luowei} \sur{Zhou}}\email{luozhou@microsoft.com}
\author[3]{\fnm{Steven J.} \sur{Yule}}\email{steven.yule@ed.ac.uk}
\author[4]{\fnm{Roger D.} \sur{Dias}}\email{rdias@bwh.harvard.edu}
\author[5]{\fnm{Milisa} \sur{Manojlovich}}\email{mmanojlo@med.umich.edu}
\author[6]{\fnm{Francis D.} \sur{Pagani}} \email{fpagani@med.umich.edu}
\author[6]{\fnm{Donald S.} \sur{Likosky}}\email{likosky@med.umich.edu}
\author*[1]{\fnm{Jason J.} \sur{Corso}}\email{jjcorso@umich.edu}

\affil*[1]{\orgdiv{EECS}, \orgname{University of Michigan}, \orgaddress{\city{Ann Arbor}, \state{MI}}}

\affil[2]{\orgdiv{Cloud and AI}, \orgname{Microsoft}, \orgaddress{\city{Redmond}, \state{WA}}}

\affil[3]{\orgdiv{Clinical Surgery}, \orgname{University of Edinburgh}, \orgaddress{\city{Edinburgh}, \country{SCT UK}}}

\affil[4]{\orgdiv{Emergency Medicine}, \orgname{Harvard Medical School}, \orgaddress{\city{Boston}, \state{MA}}}

\affil[5]{\orgdiv{School of Nursing}, \orgname{University of Michigan}, \orgaddress{\city{Ann Arbor}, \state{MI}}}

\affil[6]{\orgdiv{Cardiac Surgery}, \orgname{University of Michigan}, \orgaddress{\city{Ann Arbor}, \state{MI}}}


\abstract{
\subsubsection*{Purpose}
 Articulated hand pose tracking is an under-explored problem that carries the potential for use in an extensive number of applications, especially in the medical domain. With a robust and accurate tracking system on surgical videos, the motion dynamics and movement patterns of the hands can be captured and analyzed for many rich tasks. 
\subsubsection*{Methods}
In this work, we propose a novel hand pose estimation model, \ourdnn,
which improves detection and tracking accuracy by incorporating a pose prior into its prediction. We show improvements over state-of-the-art methods which provide frame-wise independent predictions, by following a temporally guided approach that effectively leverages past predictions.
\subsubsection*{Results}
We collect \ourdataset, the first dataset that provides multi-instance articulated hand pose annotations for videos.
Our dataset provides over \totalanno\ annotated hand poses from publicly available surgical videos and bounding boxes, pose annotations, and tracking IDs to enable multi-instance tracking.
When evaluated on \ourdataset, we show our method outperforms the state-of-the-art approach using mean Average Precision (mAP), to measure pose estimation accuracy, and Multiple Object Tracking Accuracy (MOTA), to assess pose tracking performance. 
\subsubsection*{Conclusion}
In comparison to a frame-wise independent strategy, we show greater performance in detecting and tracking hand poses and more substantial impact on localization accuracy.
This has positive implications in generating more accurate representations of hands in the scene to be used for targeted downstream tasks.
}

\keywords{Articulated Pose, Surgical Videos, Computer Vision, Hand Pose, Video Tracking}



\maketitle
\section{Introduction}
\label{sec:introduction}

Machine learning and computer vision have become increasingly integrated with healthcare in the medical community. This is apparent in the myriad of tasks such as 
tumor segmentation \cite{malathi2019brain}, technical skill assessment \cite{dias2019using, tao2012sparse, zappella2013surgical, forestier2018surgical, kumar2015surgical}, and tool detection and tracking \cite{sarikaya2017detection, colleoni2019deep, ni2019rasnet, nwoye2019weakly}.
Here we study the problem of articulated hand pose tracking in the surgical domain.
Tracking hand poses can facilitate other useful tasks such as technical skill assessment, temporal action recognition, and training surgical residents.
Pose tracking in the computer vision community is primarily centered around human poses  \cite{andriluka2018posetrack, xiao2018simple, bertasius2019learning, sun2019deep, ning2020lighttrack, wang2020combining, cao2017realtime, raaj2019efficient, jin2019multi}, while medical works focus on detecting and tracking surgical instruments \cite{sarikaya2017detection, colleoni2019deep, ni2019rasnet, nwoye2019weakly}.
Tracking surgical instruments is useful but these instruments are inherent to the surgical procedures seen during training.
Instead we abstract away the emphasis on surgical instruments
where articulated hand tracking will be more applicable to broad surgical tasks. Articulated hand pose tracking can highlight important properties such as grip, motion, and tension that human experts often attend to when evaluating videos.
 
\begin{figure}[t]
    \centering
    \includegraphics[width=0.85\columnwidth]{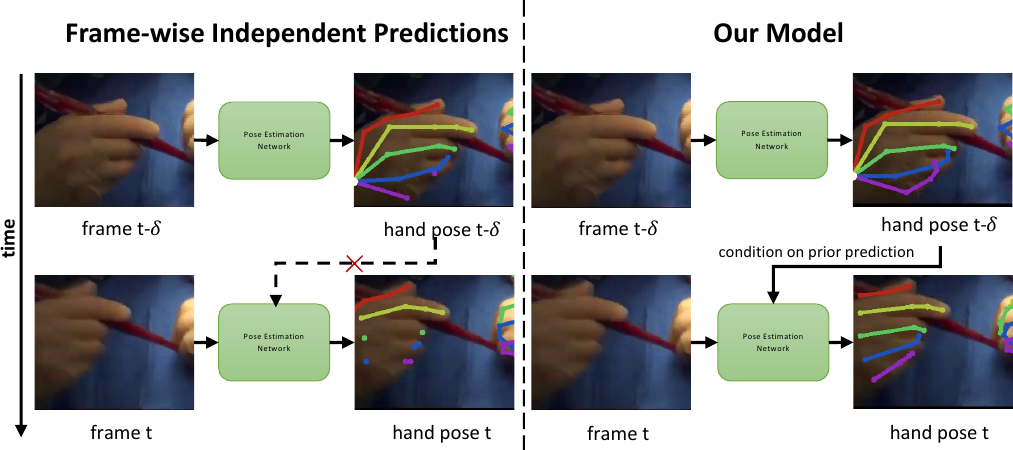}
    \caption{On the left, a method only performing frame-wise independent predictions may miss out on properly localizing joints, while on the right, temporally passing past predictions from previous frames improves the network's localization.}
    \label{fig:teaser}
\end{figure}

A challenge in pose tracking is the temporal consistency of predictions between frames, the lack of which leads to flickering and improbable changes in estimated poses.
Existing works \cite{andriluka2018posetrack, sun2019deep, cao2017realtime, raaj2019efficient, jin2019multi} in articulated pose tracking use frame-wise independent predictions along with post-processing when tracking \cite{xiao2018simple, bertasius2019learning, wang2020combining, ning2020lighttrack} to gather temporal context.
However, they do not integrate past inferences when localizing joints. We address this by proposing \ourdnn, a new model that performs predictions conditioned on the pose estimates from prior frames.
In Fig. \ref{fig:teaser}, we show a comparison of both approaches: the baseline using frame-wise independent predictions and our model using conditional predictions. The initial estimate may fluctuate due to varying factors such as lighting, hand orientation, or motion blur. But we find that using prior predictions as guidance, we can improve our localization accuracy.
The internal representation of this object's state (position, appearance, and classification) is a function of its current state and previous states.
By learning this Markovian prior for the prediction of hand joints, we can improve both pose estimation and consequently tracking accuracy.

There is a lack of data and benchmarks for articulated hand pose tracking.
To address this, we collect a novel dataset featuring intra-operative videos of real surgeries, \ourdataset. We annotate the articulated hand poses of surgeons which subsumes both surgical instrument and non-instrument actions, e.g. suturing, knot-tying, and gesturing. 
We are, to the best of our knowledge, the first to introduce a labeled dataset for both detection and tracking of multiple articulated hand poses.
We benchmark our dataset against existing tracking baselines and demonstrate the superiority of our proposed approach on both hand pose estimation and tracking.




Our contributions are as follows:
\begin{itemize}
    \item We introduce \ourdnn, a novel deep network that takes advantage of confident prior predictions to improve localization accuracy and tracking consistency. 
    \item We present \ourdataset \footnote{Both the code and dataset are available at \url{https://github.com/MichiganCOG/Surgical_Hands_RELEASE}}, a new video dataset for multi-instance articulated hand pose estimation and tracking in the surgical domain.
    \item We set new state-of-the-art benchmark performance on \ourdataset.
    
\end{itemize}

\section{Related Works}
\label{sec:related_works}

\subsection{Articulated Pose Estimation and Tracking}

\subsubsection{Surgical Instruments}
Data-driven methods in the medical video domain primarily involve RAS videos. Works in this space \cite{tao2012sparse, zappella2013surgical, forestier2018surgical} traditionally use kinematic data directly, requiring an external apparatus to capture these measurements. But full kinematic information is only available for robotic-controlled tools, even less so for hand-held instruments.
Adding any external apparatus to capture kinematic data can negatively impact the costs, flexibility, and performance of certain operations.
For detection, pure computer vision-based approaches extract information directly from video data to perform object detection.
Many vision works use a region proposal network to perform localization \cite{khalid2020evaluation, jin2018tool, sarikaya2017detection}, segmentation \cite{laina2017concurrent, ni2019rasnet}, and articulated pose estimation \cite{du2018articulated, colleoni2019deep} from images. 

To incorporate tracking, existing works may use a similarity function based on weighted mutual information \cite{richa2011visual} or Bayesian filtering as part of a minimization problem \cite{sznitman2012unified}. 
Nwoye et al. \cite{nwoye2019weakly} are the first to measure the Multiple Object Tracking Accuracy (MOTA) \cite{bernardin2008evaluating} for surgical instruments in this setting, using a weakly-supervised approach with coarse binary labels indicating the presence or absence of seven surgical instruments.
However, their evaluation contains at most one unique type of tool at each frame; hence, can be narrowed down to an object detection problem. Unlike their work, we track multiple instances of the same object in each frame.
We also use MOTA as part of our benchmark when tracking hands in our videos.

\begin{figure}[t]
    \centering
    \includegraphics[width=\columnwidth]{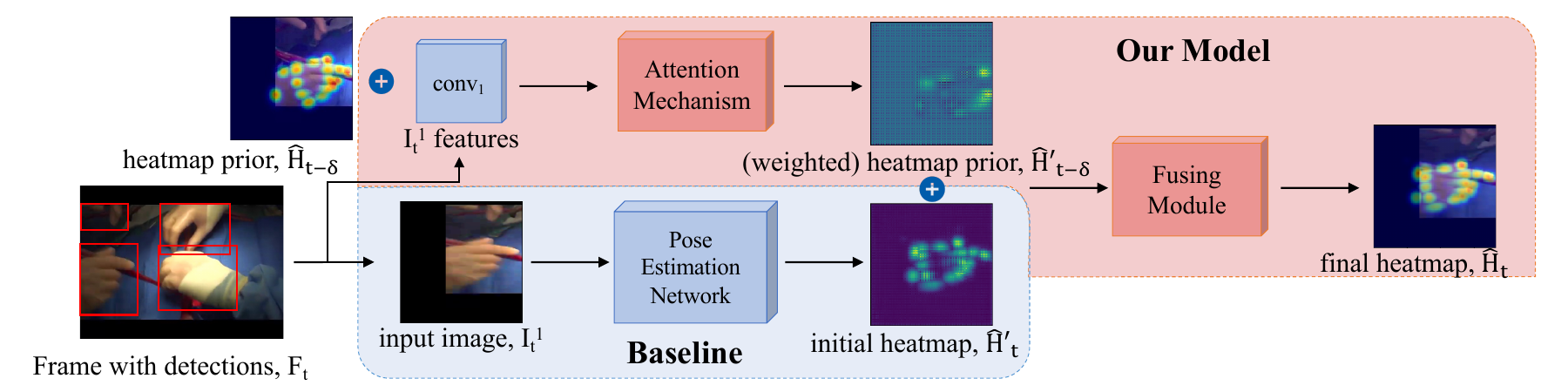}
    \caption{The baseline generates a heatmap, $\hat{\mathcal{H}'_t}$, for each detection using a pose estimation network. In our model, we provide additional information by incorporating a heatmap prior from $t - \delta$. Concatenating the image features at $t$ with $\hat{\mathcal{H}}_{t-\delta}$, we pass this through our attention mechanism to produce a weighted heatmap prior, $\hat{\mathcal{H}}'_{t-\delta}$. Both $\hat{\mathcal{H}'_t}$ and $\hat{\mathcal{H}}'_{t-\delta}$ are concatenated and passed through the fusing module, using context from both heatmaps to produce the final articulated hand pose. (The initial and final heatmaps represent real outputs from the network, while the heatmap prior (during training) shows ground truth at $t - \delta$)}
    \label{fig:overview}
\end{figure}

\subsubsection{Human Pose}
Pose estimation and tracking is commonly applied to images and videos of people, grouped into top-down \cite{xiao2018simple, bertasius2019learning, sun2019deep, ning2020lighttrack, wang2020combining} and bottom-up \cite{cao2017realtime, raaj2019efficient, jin2019multi} strategies.
Top-down methods detect all persons from an image, then regress each human pose independently using a pose estimation network.
Bottom-up methods detect all joints in an image, and use bipartite matching and graph minimization techniques to assign joints to each person.
As top-down approaches typically perform best in practice, we follow this paradigm.
For tracking, \cite{xiao2018simple} uses a greedy matching from IoU (intersection-over-union) overlap and optical flow to propagate bounding boxes between frames, \cite{bertasius2019learning} use deformable convolutions to warp predictions between frames, and \cite{ning2020lighttrack} introduce a Graph Convolutional Network (GCN) \cite{kipf2017semi} to match learned embeddings between human poses.
A GCN is a neural network whose input consists of a set of nodes and edges, performing convolution operations on the relations of nodes. The inherent structure of this graph can improve quality of learned features as well as abstracting from limitations of a 2D space. 
These approaches spatially shift pose predictions, which cannot overcome certain factors (e.g. missed detections). In contrast, we address this problem at the detection step by integrating past pose observation(s) into each new predicted output.

\subsubsection{Hand Pose}
Current works on 2D hand pose estimation \cite{simon2017hand, santavas2020attention, zimmermann2019freihand} are analogous to human pose estimation. Zhang et al. \cite{zhang2017hand} performs pose tracking, using a disparity map from stereo camera inputs to estimate a 3D hand pose. However their data consists of only a single subject's hand and at most one detection per frame.
There are many image datasets \cite{simon2017hand, zhang2017hand, gomez2019large, zimmermann2019freihand} for hand pose estimation, from a combination of manual, synthetic, and predicted annotations. But none satisfy the conditions of multiple object instances and tracking from video, more so in a surgical setting.
Therefore, we introduce the \ourdataset\ dataset for multi-instance articulated hand pose tracking.
Our dataset includes varying lighting conditions, fast movement, and diversity in scene appearances. Distinctively, we also include gloved hands, which appear in contrasting colors such as latex and green.  

\section{Method}
\label{sec:method}
We propose \ourdnn, to perform articulated pose detection and tracking by incorporating previous observations as prior guidance.
We show our model in Fig. \ref{fig:overview}. 
While the baseline produces a heatmap from each hand using a pose estimation network, we leverage past predictions to produce conditioned hand pose outputs, improving detection performance during inference.
While we design \ourdnn\ with video data in mind, we begin with pretraining on image data, finetuning on our video dataset, \ourdataset, and lastly, comparing between different tracking methods.

\subsection{Hand Pose Estimation in Images}
\label{sec:hand_pose_images}
We first pretrain on image data, defining the input and output for the pose estimation network, $P$, as $\hat{\mathcal{H}} = P(\mathcal{I})$.
The input is an image crop $\mathcal{I}$, $\mathcal{I} \in \mathbb{R}^{H \times W \times 3}$, and the output is a predicted heatmap $\hat{\mathcal{H}}$, $\hat{\mathcal{H}} \in \mathbb{R}^{H' \times W' \times J}$. Here $H, W$ represents the input image height and width and $H', W'$ are the output heatmap height and widths. $J$ represents the number of predicted joints of each hand. 
Each image crop is scaled to $2.2$ times the total area of the hand bounding box.
We train using the mean squared error (MSE) between the ground truth and predicted heatmaps as 
$\mathcal{L} = \Vert (\mathcal{H} - \hat{\mathcal{H}}) \odot \mathcal{M} \Vert^2$.
The ground truth heatmaps, $\mathcal{H}$, are generated from 2D Gaussians centered on each annotated keypoint.
$\mathcal{M}$, is included to mask out un-annotated joints.
The output joint locations are the max value positions in the third channel of $\hat{\mathcal{H}}$.
After pretraining, we finetune our model on videos to learn conditional hand pose predictions.

\begin{figure*}[t]
    \centering
    \includegraphics[width=0.85\columnwidth]{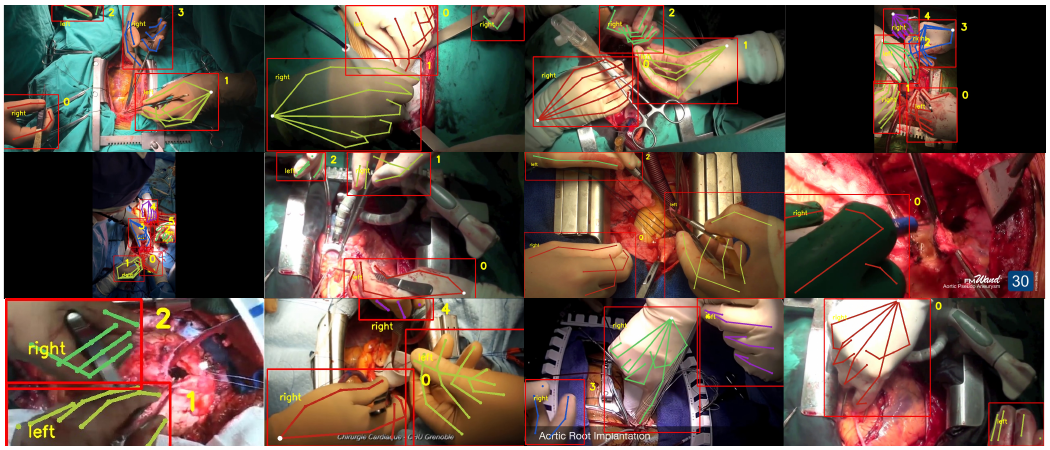}
    \caption{We show samples from our annotations. Each hand is labeled with a bounding box, handedness, tracking id, and visibility of joints.}
    \label{fig:annotation_samples}
\end{figure*}

\subsection{Hand Pose Estimation in Videos} \label{sec:hand_pose_recursive}
While image data cannot be used to learn our conditional hand pose predictions, we can initialize weights to speed up our training process and improve generalizability.
We finetune \ourdnn\ on \ourdataset, shown in the top portion of Fig. \ref{fig:overview}.
To incorporate a prior branch, we introduce a heatmap prior, $\hat{\mathcal{H}}_{t-\delta}$, a pose estimate of the same object from $t-\delta$. Our model performs conditional predictions, defined as
\begin{align}
    \hat{\mathcal{H}}_t = M_{fus}(P (\mathcal{I}_t); M_{att}(v_t; \hat{\mathcal{H}}_{t-\delta})) \label{eq:recursive_model}.
\end{align}
In contrast to our previous definition of $P$, $\hat{\mathcal{H}}_t$ is now conditioned on predictions at a previous time step $t - \delta$.
Our model is further composed of two branches: the attention mechanism, $M_{att}$, and the fusing module, $M_{fus}$.
$M_{att}$ contextualizes the prior heatmap prediction, $\hat{\mathcal{H}}_{t-\delta}$, with image features, $v_t$ ($conv\_1$ in our experiments), at time $t$. This branch relates the visual representation and the localized heatmap prior, ideally learning to weight each joint prior accordingly.
$M_{fus}$ produces a merged final heatmap from the initial prediction, $\hat{\mathcal{H}}'_t$, and weighted heatmap prior, $\hat{\mathcal{H}}'_{t-\delta}$.
$M_{att}$ and $M_{fus}$ are both composed of two convolutional layers, followed by transposed convolution, with ReLU non-linearities in-between. 

During training the prior is selected from frame $t - \delta$. If the object does not exist at that frame, we use earlier frames up until the first occurrence. If a corresponding object does not exist on any previous frames, then the prior, $\hat{\mathcal{H}}_{t-\delta}$, is set as a zeros heatmap. 
This is expected behavior during evaluation, because priors do not yet exist at frame one.
Also during evaluation, unlike training, the prior associated with the current detection is unknown. Given $n$ priors from time $t-1$, $\{ \hat{\mathcal{H}}_{t-1}^{1}, \hat{\mathcal{H}}_{t-1}^{2}, \dots \hat{\mathcal{H}}_{t-1}^{n}\}$, and $k$ detections at time $t$, $\{ \hat{\mathcal{I}}_{t-1}^{1}, \hat{\mathcal{I}}_{t-1}^{2}, \dots \hat{\mathcal{I}}_{t-1}^{k}\}$ we pass all pairs through the network to generate candidates. The heatmap with the highest average confidence score is selected as the output for that detection.

\subsection{Matching Strategies for Tracking} \label{sec:graph_pose_track}

Following the detect-then-track paradigm, we require a matching strategy to performing tracking. Given $n$ hands at time $t-1$ and $m$ hands at time $t$ we use a similarity function to derive similarity measures between each pair at $t-1$ and $t$.
Common methods are intersection-over-union (IoU) of bounding boxes, average L2-distance of the predicted joint locations, or L2-distance between the graph pose embeddings.
Similar to Ning et al. \cite{ning2020lighttrack} we train a GCN to output the embedding of each input hand pose, $\mathcal{X}$, defined simply as $\hat{p} = GCN(\mathcal{X})$.
Here $\mathcal{X} \in \mathbb{R}^{J \times C}$, where $J$ is the number of joints and $C$ is the number of channels. For training, we use the contrastive loss \cite{hadsell2006dimensionality},
$\mathcal{L} = \frac{1}{2} \left( y * d + (1 - y) * max \left( 0, (m - d)^2 \right)\right)$.
The contrastive loss places embeddings close in perceptual distance.
For a pair of embeddings $\hat{p}_v^1$ and $\hat{p}_v^2$, the variable $d$ represents the L2-distance between the two, $d = \Vert \hat{p}_v^1 - \hat{p}_v^2\Vert^2$. $y$ is a binary label indicating the same hand, $1$, or different hands, $0$. $m$ is the margin variable, a hyperparameter used for tuning.
For each item in our minibatch, positive pairs are selected between adjacent frames with probability $p=0.5$ and negative pairs are selected from the same video with $p=0.4$ or from a different video with $p=0.1$. 
We evaluate our trained GCN models using the classification accuracy between pairs of selected hands, achieving classification accuracies of $>97\%$.

\section{Dataset}
\label{sec:dataset}

We lack data for training and benchmarking models on multi-instance hand tracking. Therefore we introduce \ourdataset, a novel video dataset for multi-instance articulated hand pose estimation and tracking in the surgical domain, the first of its kind.
From publicly available data, we collect $\totalvids$ videos with a view of the hands of surgical team members during the operation.
From those videos, we extract $\totalclips$ clips sampled at $8$ frames per second and collect bounding box, class label, tracking id, and pose annotations using Amazon Mechanical Turk (AMT) and a modified version of Visipedia Annotation Tools \footnote{\url{https://github.com/visipedia/annotation_tools}}. We show samples of our annotations in Fig. \ref{fig:annotation_samples}.
Each hand is labeled with the handedness (left/right), 21 joints, and properties for each joint: visible, occluded or non-available.
Visible implies that the joint is visibly on screen, occluded means the joint is obstructed but its position can be estimated, not-available means the joint position cannot be inferred or it is off-screen. From our collected data, we have a total $2,838$ annotated frames and $\totalannotations$ unique hand annotations from 21 unique annotators. Each annotated frame contains a mean of $2.88$ hands, median of $3$ hands, and a maximum of $7$ hands. 

\section{Experiments and Evaluation}
\label{sec:experiments}

\subsection{Implementation Details}
We adopt a ResNet-152 pose estimation model \cite{xiao2018simple} to first train on hand pose image data, CMU Manual Hands and Synthetic Hands \cite{simon2017hand}.
We use a batch size of $16$, training for $30$ epochs, with an Adam optimizer and a learning rate of $1e^{-3}$.
When finetuning on \ourdataset\, we use leave-one-out cross-validation and split our data into $\totalvids$ different folds. Clips belonging to the same video are in the same validation fold, and the reported metrics are averaged across all folds.
We employ a variant of curriculum learning that gradually transitions to predicted priors from ground truth priors. A predicted prior at $t-\delta$ is sampled with a probability of $p = 0.10 * epoch$, until only predictions are used for training at epoch $10$ and onward. We empirically select $\delta=3$ during training.
For all training, we apply random rotations and horizontal flipping as data augmentation.
When training the GCN for tracking, we using a batch size of $32$ and train for $60$ epochs and an initial learning rate of $1e^{-3}$. We normalize $\mathcal{X}$ to 0-1, relative to keypoint positions along the bounding box. 
The input dimension for each input is $J \times C$ where $J$ represents the number of joints and $C$ is the number of channels. We use $C=2$ for x-y coordinates and $C=3$ to include annotation state (0 = unannotated, 1 = annotated, or 0-1 for predicted keypoints). We adopt
a two-layer Spatio-Temporal GCN \cite{yan2018spatial, ning2020lighttrack} to output a 128-dimensional embedding of each pose.

\subsection{Detection Performance}
\begin{figure*}[t]
    \centering
    \includegraphics[width=0.85\textwidth]{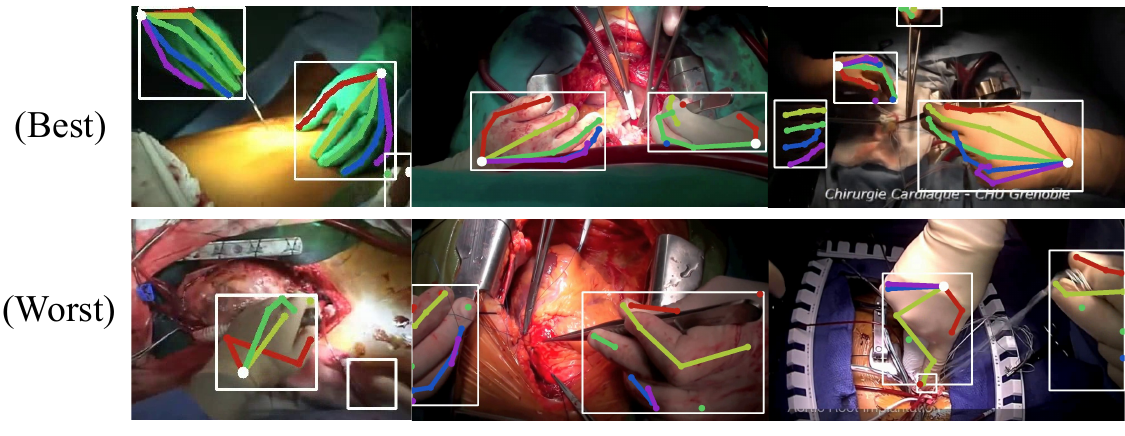}
    \caption{We show qualitative samples of frames from the best performing (top row) and lower performing (bottom row) videos. (Best viewed in color). }
    \label{fig:qualitative_examples_det}
\end{figure*}

We evaluate detection performance using mean Average Precision (mAP), the choice metric in human pose evaluation, on our \ourdataset\ dataset. MAP is computed using the Probability of Correct Keypoints (PCK), measuring the probability of correctly localizing keypoints within a normalized threshold distance, $\sigma$.
%
This threshold distance, $\sigma$=0.2, is empirically chosen to be roughly the ratio between the length of a thumb joint and the enclosing bounding box. Pose predictions are matched to ground truth poses based on the highest PCK and unassigned predictions are counted as false positives. AP for each joint is computed and mAP is reported across the entire dataset.
In Table \ref{tab:ap} we report the mAP at the highest MOTA score (defined in the next section) for each model.
With our recursive heatmap strategy we are able to obtain higher average precision across the different joints in the hand.
In Fig. \ref{fig:qualitative_examples_det} we show qualitative examples of our hand pose estimation on various frames from our \ourdataset\ dataset. The top row clips are sampled from the best performing clips, while the bottom row are from the worst performing clips.
We see that the model suffers most in cases of heavy occlusion, where the camera view excludes the majority of the hand. Ambiguity in the position of the hand furthers the localization errors, e.g. top-down view with most fingers occluded. The best performing cases are those with balanced lighting and an unambiguous view of the first few digits.

\begin{table}[t]
    \centering
    \caption{Mean Average Precision (mAP). Performance is averaged across all folds} \label{tab:ap}
        \begin{tabular}{l l l l l l l l}
        \toprule 
          Model &Wrist &Thumb &Index &Middle &Ring &Pinky &mAP \\ \midrule
            Baseline \cite{xiao2018simple} & \textbf{67.23} &60.12 &63.29 &53.77 &48.29 &39.28 &53.59 \\
            Our model & 65.51 &\textbf{62.66} &\textbf{64.99} &\textbf{57.88} &\textbf{51.40} &\textbf{44.26} &\textbf{56.66} \\
        \bottomrule
    \end{tabular}
\end{table}

\begin{table*}[t]
    \centering
    \caption{We optimize for the Multiple Object Tracking Accuracy (MOTA), each performance metric is averaged across all validation folds} \label{tab:mota}
    \resizebox{\columnwidth}{!}{
        \begin{tabular}{l l l l l l l l l l l l}
            \toprule
              Model &MOTA &MOTA &MOTA &MOTA &MOTA &MOTA &MOTA &MOTP &Prec. &Rec. & F$_1$ Score \\
            &Wrist &Thumb &Index &Middle &Ring &Pinky &Total &Total &Total &Total& Total \\ \midrule
            Baseline \cite{xiao2018simple} &\textbf{36.7} &\textbf{45.83} &57.35 &45.53 &34.63 & 8.49 &38.27 &85 &\textbf{78.3} &59.13 & 67.37\\
            Our Model &30.99 &44.74 &\textbf{58.21} &\textbf{48.90} &\textbf{36.46} & \textbf{10.39} &\textbf{39.31} &\textbf{85.28} &77.61 &\textbf{62.69}& \textbf{69.35}\\
            \bottomrule
        \end{tabular}
        }
\end{table*}

\begin{figure*}[t]
    \centering
    \includegraphics[width=\textwidth]{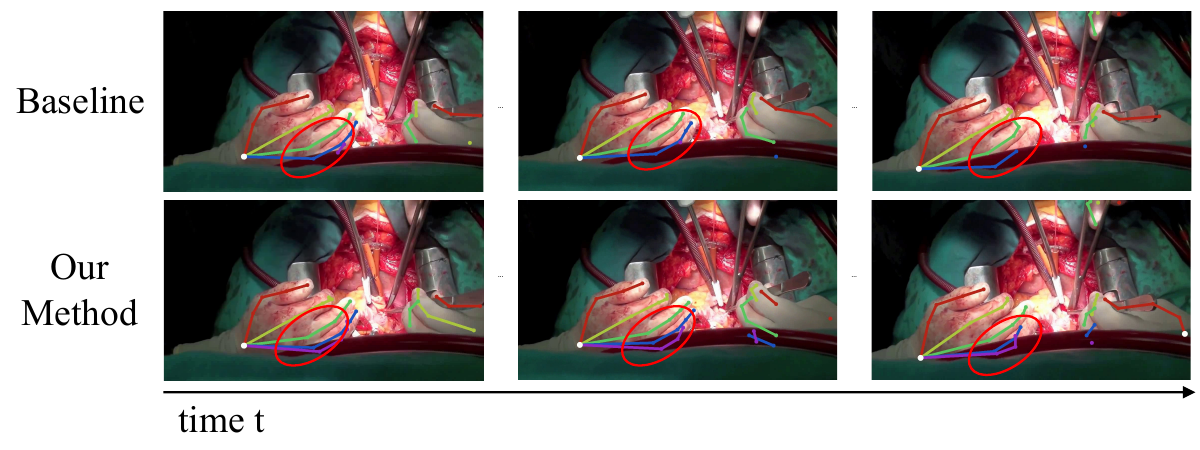}
    \caption{We show a qualitative comparison between the baseline model and our method. We note a higher recall and consistency between frames, as shown for the hand to the left. Even when the pinky finger is not visible, the past predictions reinforces those joint locations.}
    \label{fig:qualitative_comparisons}
\end{figure*}

\begin{table}[t]
\caption{MOTA performance between matching strategies, averaged across all folds. Each row is optimized for highest MOTA performance. Matching strategies share the same base model, so it is possible for them to share the same mAP score.}\label{tab:matching_strategy}
\begin{center}
    \begin{tabular}{l l l l l l }
    \toprule 
    & &\multicolumn{2}{c}{Perfect Det.} &\multicolumn{2}{c}{Object Det.} \\
    Model &Matching Strategy &mAP &MOTA &mAP &MOTA \\ \midrule
    \multirow{3}{*}{Baseline \cite{xiao2018simple}} &IoU & 53.59 & 38.27 & 48.15 & 31.46 \\
    &L2 & 52.65 & 37.78 & 47.44 & 31.14 \\
    &GCN & 52.65 & 36.78 & 47.44 & 30.03 \\\midrule 
    \multirow{3}{*}{Our Model} &IoU &\textbf{56.66} &\textbf{39.31} & \textbf{50.04} &\textbf{33.19} \\
    &L2 & 56.66 & 38.94 & 50.04 & 32.84 \\
    &GCN & 56.66 & 38.22 & 50.04 & 32.24 \\
    \bottomrule 
    \end{tabular}
\end{center}
\end{table}

\subsection{Tracking Performance}
To measure tracking performance, we use Multiple Object Tracking Accuracy (MOTA) which also takes into account the consistency of localized keypoints between frames.
MOTA \cite{bernardin2008evaluating} is defined as:
\begin{align}
MOTA = 1 - \frac{\sum_t \left( FN_t + FP_t + IDSW_t \right)}{\sum_t G_t}
\label{eq:mota}    
\end{align}
This encapsulates errors that may occur during multiple object tracking: false negatives ($FN$), false positives ($FP$), and identity switches ($IDSW$). 
$FN$ are joints for which no hypothesis/prediction was given, $FP$ are the hypothesis for which no real joints exists, and $IDSW$ are occurrences where the tracking id for two joints are swapped. $G$ represents the total number of ground truth joints. The range of values for the $MOTA$ score is $(- \infty$ to $100]$.

We measure perform tracking using three methods: IoU, L2-distance, and GCN. Intersection-over-union (IoU) 
measures overlap of two bounding boxes using the ratio: area of intersection over total area, between subsequent frames in our case.
L2-distance measures the average L2 distance of regressed keypoints between frames.
GCN measures the embedding similarity between the encoded keypoints to determine matches.
We show quantitative results from our experiments in Table~\ref{tab:matching_strategy} and the per-joint performance in Table~\ref{tab:mota}.
Each row is maximized for the highest MOTA score across all hyperparameters, shown along with its corresponding mAP.
Our method has a higher MOTA score across all of the videos, but our corresponding mAP scores are greater by a much larger margin. This points to our advantage from temporally leveraging predictions from previous frames during the detection step.
We show an example in Fig. \ref{fig:qualitative_comparisons}, in a frame-by-frame comparison between the baseline and our method, we note a higher recall and improved localization. 
While the last digit is obstructed, its position can be reasonably inferred.
In the last two columns of Table~\ref{tab:mota} we use an object detector to detect hands, the prior two columns (perfect detections) use the manual annotations. Training an object detector on 100 Days of Hands (100DOH) \cite{shan2020understanding}, we see a lower localization and tracking accuracy
but a consistent trend from the baseline. The quality of the detections serve as a bottleneck, but the margins of improvements are very similar. While trained with perfect detections as priors, they are not required to maintain performance in practice.

\subsection{Ablation Analysis}
We perform an ablative analysis on the convolutional map in $M_{att}$ and the fusing module $M_{fus}$. We experiment with no prior convolutional feature map (NC), no attention mechanism (NA), and removal of both (NC-NA), showing our results in Table \ref{tab:ablation_table}.
Our full model has the highest scores overall.
The attention mechanism and convolutional feature maps have opposing effects on the mAP and MOTA scores.
The NC model does not use a convolutional feature map from frame $t$, so the fusing module is applied directly to both un-altered heatmaps from $t-\delta$ and $t$. We found this increases the mAP value, but lowers the MOTA score.
The NA model directly concatenates the convolutional features and the heatmaps, with no attention mechanism. This has the opposite effect, decreasing the mAP significantly but slightly increasing the overall MOTA score.
Without contextual convolutional features (NC and NC-NA), the model can still learn to use the prior prediction and improve its detection score. On the contrary, no attention mechanism brings a drop in mAP, which may be attributed to an unrefined prior with noisy features.
The small increase in the MOTA score is likely from fewer false positives produced by that model, due to a slightly lower mAP.

We also explore the value of our hyperparameter, $\delta$, during training. We use values $\delta=\{1,2,3,4\}$ and show our results in Table \ref{tab:delta}.
Optimizing for highest MOTA score, we found $\delta=3$ to be best with $39.31$, followed by $\delta=1$ with a smaller MOTA score (39.03) but a higher mAP ($58.64$ vs $56.66$).
We find a non-linear correlation between the mAP and MOTA scores, showing a trade-off in mAP when optimizing for the tracking performance.
The best strategy is one that maximizes MOTA accuracy with minimal loss in localization precision.

\begin{table}[t]
\caption{Ablation analysis using IoU matching strategy ($\delta=1$). NC = No convolutional feature map, NA = No attention mechanism.}\label{tab:ablation_table}
\begin{center}
    \begin{tabular}{l l l l}
    \toprule 
    & &\multicolumn{2}{c}{Perfect Det.}\\
    Model Variant &Matching Strategy &mAP &MOTA\\ \midrule
    NC-NA & IoU & 55.23 & 38.31 \\
    NC & IoU & 56.00 & 38.13 \\
    NA & IoU & 54.70 & 38.45 \\
    Full model & IoU & \textbf{56.66} & \textbf{39.31} \\ 
    \bottomrule 
    \end{tabular}
\end{center}
\end{table}

\begin{table}[t]
\caption{Effect of $\delta$. Each model is trained with a separate $\delta$ value} \label{tab:delta}
\begin{center}
    \begin{tabular}{l l l l}
    \toprule 
    & &\multicolumn{2}{c}{Perfect Det.}\\
    Model Variant & Matching Strategy & mAP & MOTA \\ \midrule
         $\delta = 1$ & IoU & \textbf{58.64} & 39.03\\
         $\delta = 2$ & IoU & 54.71 & 38.42\\
         $\delta = 3$ & IoU & 56.66 & \textbf{39.31}\\
         $\delta = 4$ & IoU & 56.35 & 38.09\\
    \bottomrule
    \end{tabular}
\end{center}
\end{table}

\subsection{Evaluation on Human Pose} \label{sec:human_pose}
\begin{figure}[t]
    \centering
    \includegraphics[width=0.9\columnwidth]{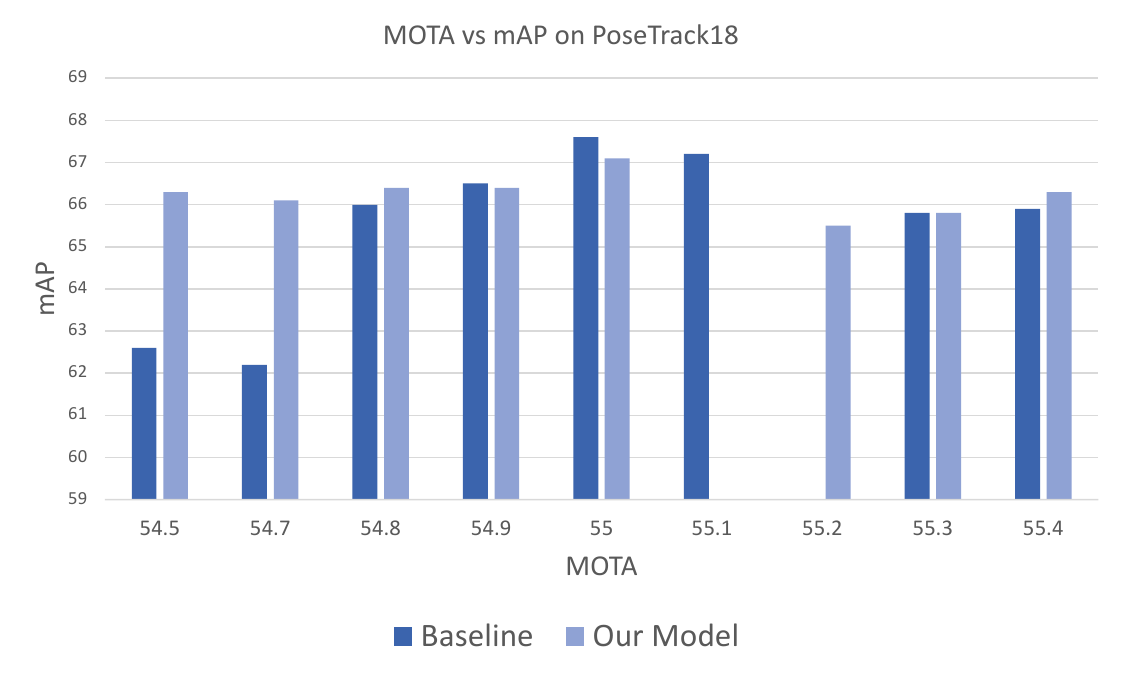}
    \caption{Optimized for maximum MOTA score, we show the top performing models on PoseTrack18.  Consistent with our earlier findings, our model maintains a higher mAP for comparable MOTA scores.}
    \label{fig:posetrack_exp}
\end{figure}

We executed additional experiments on the PoseTrack18 dataset between our model and our re-implementation of the baseline.
From Fig. \ref{fig:posetrack_exp}, we show a narrowed gap in performance but our findings are consistent with our earlier experiments.
Our model maintains a higher mAP score for the highest MOTA values.
Given the trade-off that occurs between mAP and MOTA, this means our model is more likely to retain its localization precision at higher tracking accuracies.

\section{Conclusion} \label{sec:conclusion}

In this work, we introduce \ourdataset, the first articulated multi-hand pose tracking dataset of its kind. Additionally we introduce \ourdnn, a novel network that makes conditional hand pose predictions by incorporating past observations as priors.
We show that when compared with a frame-wise independent strategy, we have better performance in localizing and tracking hand poses.
More so, a higher localization accuracy for comparable tracking performance.
While tracking drives the consistency of joints through time, the actual shape and characteristics of the hand is described by the localization precision. With a higher localization precision and better tracking still, we can guarantee a better representation of the hands in the scene.
%
While not the focus of this work a reliable hand tracking method can provide a salient signal that can be used to approximate surgical skill or understanding actions.

\section{Statements and Declarations}
\begin{itemize}
    \item \textbf{Funding} This project was supported by the National Heart, Lung, and Blood Institute (NHLBI: R01HL146619) and the University of Michigan (U-M's Mcubed Program). Opinions expressed in this manuscript do not represent those of The NIH or the US Department of Health and Human Services or the US Department of Veterans Affairs.
    \item \textbf{Conflict of Interest} The authors declare that they have no conflict of interest.
    \item \textbf{Ethical Approval} This article does not contain any studies with human participants or animals performed by any of the authors.
    \item \textbf{Informed Consent} This article does not contain patient data.
\end{itemize}


\bibliography{sn-bibliography}


\end{document}